\documentclass[conference]{IEEEtran}
\IEEEoverridecommandlockouts
\usepackage{cite}
\usepackage{amsmath,amssymb,amsfonts}
\usepackage{algorithmic}
\usepackage{graphicx}
\usepackage{textcomp}
\usepackage{xcolor}
\def\BibTeX{{\rm B\kern-.05em{\sc i\kern-.025em b}\kern-.08em
    T\kern-.1667em\lower.7ex\hbox{E}\kern-.125emX}}
\usepackage{url}

\usepackage{graphicx}
\usepackage{booktabs}
\usepackage{color}
\usepackage{amssymb}
\usepackage{makecell}
\usepackage{bbm}
\usepackage{caption}
\usepackage{url}

\usepackage{amsmath}
\usepackage{amsfonts}
\usepackage[export]{adjustbox}

\usepackage[inline]{enumitem}

\newif\ifdraft
\drafttrue

\usepackage{multirow}

\begin{document}
\title{Messing Up 3D Virtual Environments: \\ Transferable Adversarial 3D Objects} 

\author{\IEEEauthorblockN{\hskip 16mm Enrico Meloni$^{1,2}$, Matteo Tiezzi$^1$,}
\IEEEauthorblockA{\textit{$^1$Dept. of Information Engineering and Mathematics} \\
\textit{University of Siena}\\
Siena, Italy}
\and
\IEEEauthorblockN{\hskip -14mm Luca Pasqualini$^1$, Marco Gori$^{1,3}$,}
\IEEEauthorblockA{\textit{$^2$Dept. of Information Engineering} \\
\textit{University of Florence}\\
Florence, Italy}
\and
\IEEEauthorblockN{\hskip -30mm Stefano Melacci$^1$}
\IEEEauthorblockA{\textit{$^3$MAASAI} \\
\textit{Universit\`{e} C\^{o}te d'Azur}\\
Nice, France \\
\hskip -14cm \{meloni,mtiezzi,pasqualini,marco,mela\}@diism.unisi.it}
}

\maketitle

\begin{abstract}
In the last few years, the scientific community showed a remarkable and increasing interest towards 3D Virtual Environments, training and testing Machine Learning-based models in realistic virtual worlds. 
On one hand, these environments could also become a mean to study the weaknesses of Machine Learning algorithms, or to simulate training settings that allow Machine Learning models to gain robustness to 3D adversarial attacks. On the other hand, their growing popularity might also attract those that aim at creating adversarial conditions to invalidate the benchmarking process, especially in the case of public environments that allow the contribution from a large community of people. Most of the existing Adversarial Machine Learning approaches are focused on static images, and little work has been done in studying how to deal with 3D environments and how a 3D object should be altered to fool a classifier that observes it. In this paper, we study how to craft adversarial 3D objects by altering their textures, using a tool chain composed of easily accessible elements. We show that it is possible, and indeed simple, to create adversarial objects using off-the-shelf limited surrogate renderers that can compute gradients with respect to the parameters of the rendering process, and, to a certain extent, to transfer the attacks to more advanced 3D engines. We propose a saliency-based attack that intersects the two classes of renderers in order to focus the alteration to those texture elements that are estimated to be effective in the target engine, evaluating its impact in popular neural classifiers.
\end{abstract}


{\let\thefootnote\relax\footnotetext{This work was partly supported by the PRIN 2017 project RexLearn, funded by the Italian Ministry of Education, University and Research (grant no. 2017TWNMH2

Accepted for publication at the IEEE International Conference on Machine Learning and Applications (ICMLA) 2021 (DOI: TBA).

\copyright\ 2021 IEEE. Personal use of this material is permitted. Permission from IEEE must be obtained for all other uses, in any current or future media, including reprinting/republishing this material for advertising or promotional purposes, creating new collective works, for resale or redistribution to servers or lists, or reuse of any copyrighted component of this work in other works.
}}

\section{Introduction}


The classic approach to the development and evaluation of Machine Learning algorithms has always relied on the availability of datasets that collect samples acquired from the real-world setting in which the algorithms are expected to operate. In the case of Computer Vision, in the last few years we observed a remarkable diffusion of simulators that are progressively conquering an important role in the development pipeline of novel algorithms \cite{beattie2016deepmind,kolve2017ai2,savva2019habitat}. The visual quality of these simulators has significantly improved, making the rendered scene very close to the real-world appearance, thus offering a manageable way to setup experiments in controlled and reproducible conditions that are visually similar to the real target setting, despite being artificially generated \cite{gan2020threedworld,Meloni2020,xia2020interactive}.

On the flip side, gaining popularity also implies that there could be a large number of people potentially eager to poison or to voluntarily corrupt data inside publicly available 3D Virtual Environments, with the aim of injecting backdoors in Machine Learning systems trained in such environments or of spoiling benchmarking procedures. As a matter of fact, once a malicious 3D object has been crafted, it can be plugged into multiple 3D scenes, spreading out its effect in an exponential manner. This might become extremely dangerous in those cases in which a large community collaborates to the development of an open project about Virtual Environments \cite{deitke2020robothor,weihs2020allenact}. Differently, when dealing with datasets of images or videos, altering some data in an adversarial manner \cite{biggio2018wild} will only affect such data, and not other images or videos that are about the same subject. Of course, moving to a more constructive perspective, it is also important to consider that purposely and admittedly augmenting 3D worlds with objects generated in an adversarial context could be useful to train more robust Machine Learning-based models or to better evaluate their quality \cite{shafahi2019adversarial}.

Even if the Adversarial Machine Learning community exploited rendered views of a 3D object to craft real objects that fool a classifier \cite{athalye2018synthesizing}, most of the efforts are oriented towards the case of datasets of images. However, in the last years a large number of novel rendering schemes were proposed, ranging from Neural Renderers \cite{eslami2018neural,sitzmann2019scene}, to more generic Differentiable Renderers \cite{kato2020differentiable}, that allow the user to compute gradients with respect to the parameters of the renderer, including meshes, textures, and others. Of course, these renderers make it easier to alter elements belonging to the 3D world with the purpose of optimizing a target objective function, thus opening for deeper investigations on how Adversarial Machine Learning can impact 3D Virtual Environments.

In this paper, we propose a novel study on the generation of adversarial 3D objects in the context of 3D Virtual Environments. 
We exploit the most straightforward differentiable rendering tools that were recently made public, directly interfaced with popular Machine Learning libraries, and that do not require robust skills in 3D graphics. Our goal is to investigate how easy is to use these tools with the precise purpose of fooling a classifier that processes views produced by a high-end 3D engine. As a matter of fact, 3D Virtual Environments are usually based on renderers (\textit{target renderers}) that have more advanced features than the ones that are supported by versatile differentiable renderers, that we will refer to as \textit{surrogate renderers}. For this reason, we investigate how effective is the process of transferring the surrogate-based adversarial 3D objects to the target renderer. We consider a realistic case study in which PyTorch3D \cite{ravi2020pytorch3d} and the popular Unity3D engine \cite{haas2014history} constitute the surrogate and target renderer, respectively. Among several 3D Virtual Environments that exploit Unity3D, we selected the recently published SAILenv \cite{Meloni2020}, that is advertised due to its simplicity in integrating it with Machine Learning algorithms. Our analysis will focus on the alteration of the textures of different objects, observed by multiple views, even if it could be extended to any other parameter supported by the surrogate renderer (mesh, lighting, etc.). 
In order to reduce the number of texture elements that are altered by the attack, we propose to consider only those texels that are estimated to be more effective when rendered by the target engine. 
The \textit{saliency} associated to different views rendered in the target engine is used as an indirect way to restrict the attack to limited regions of the textures.
We evaluate the transferability of the proposed attack strategy using popular neural classifiers, experimentally confirming that it is indeed possible to rely on simple software to setup a tool chain that can craft adversarial 3D objects for 3D Virtual Environments. 
The resulting adversarial objects can be used to augment the object library of SAILenv, and exploited by the community to improve or test the robustness of newly developed classifiers.
To the best of our knowledge, we are the first ones to provide evidence of the extent to which this process is effective, opening the road to further investigations.

\section{Background}
The three pillars that support the analysis of this paper are 3D virtual environments, rendering software, and the generation of adversarial examples, that we describe in the following subsections, together with recent related work.

\subsection{3D Virtual Environments}
\label{sec:3dv}
In recent years there has been an  emerging  paradigm  shift in the way Machine Learning algorithms are evaluated, focusing on algorithms and agents that do not simply learn from datasets of images, videos or text but instead learn in 3D Virtual Environments. Due to the improved photorealism of the rendered scenes, there has been substantial growth in the demand for 3D simulators to support a variety of research tasks \cite{beattie2016deepmind,gan2020threedworld,kolve2017ai2,Meloni2020,savva2019habitat,weihs2020allenact,xia2020interactive}. 
Most virtual environments allow for basic agent navigation in closed door environments and limited physical interaction, while some also have photo-realistic and moving objects. It not surprising to see 3D Virtual Environments dedicated to Machine Learning and, more generally, AI that are built within 3D game engines, designed to render high-quality graphics at large frame rates. Among a variety of recent 3D environments, we mention DeepMind Lab \cite{beattie2016deepmind} (Quake III Arena engine), VR Kitchen \cite{gao2019vrkitchen}, CARLA \cite{dosovitskiy2017carla} (Unreal Engine 4), AI2Thor~\cite{kolve2017ai2}, CHALET~\cite{yan2018chalet}, VirtualHome~\cite{puig2018virtualhome}, ThreeDWorld~\cite{gan2020threedworld}, SAILenv~\cite{Meloni2020} (Unity3D game engine), HabitatSim~\cite{savva2019habitat}, iGibson~\cite{xia2020interactive}, SAPIEN~\cite{xiang2020sapien} (other engines).

These environments are designed to be interfaced with high-level programming languages and, in turn, with common Machine Learning libraries. The visual quality of the rendered scenes depends both on the features of the 3D engine and on the design of the 3D models that are shared together with the environment itself. Several different tasks are studied using these 3D simulators, such as generic robot navigation, visual recognition, visual QA -- see \cite{Duan2021} and references therein. Some environments are developed in the context of open projects that might benefit from the contributions of large communities \cite{deitke2020robothor,kolve2017ai2,weihs2020allenact}, while others are based on closed source solutions \cite{gan2020threedworld}. To the best of our knowledge, their flexibility as tools to deepen the understanding of Adversarial Machine Learning has not been the subject of specific studies yet.



\subsection{Renderers}
\label{sec:rendering}
Rendering is the process of generating an image from a 3D scene by means of a computer program, which is known as the rendering software, or renderer for short. In a nutshell, and skipping several details, we could think of rendering as a function $r$ from a 3D scene $s$ and a camera $c$ to a 2D image $I_{s,c}$,
\begin{equation}
I_{s,c} = r(s, c),
\label{eq:rendering}
\end{equation}
where any $s$ is composed of 3D objects (meshes), lights, and other elements. During rendering, objects are projected onto the camera view plane, taking into account lights and the relevant properties of the objects. The nature of these properties and the influence of light vary depending on the renderer $r$ we use. 
Modern 3D engines support high-end rendering facilities, among which we mention Physically Based Rendering (PBR), 
that indicates a broader range of technologies simulating the behavior of light impacting and bouncing on the so-called materials, that allow the object to react to the light sources in a realistic manner. Each material has specific properties and texture maps defining its roughness, reflectivity, occlusion and so forth, depending on the engine specifications. 
For example, the standard shader in Unity3D~\cite{haas2014history} supports the definition of color and opacity (Albedo Texture Map) or how metallic and smooth a material should be (Metallic Smoothness Texture Map), togheter with several other properties (Albedo Map Color, Ambient Occlusion Texture Map, Smoothness Multiplier, Normal Multiplier) \cite{haas2014history}.
Differently, in non-PBR renderers most of the information commonly contained in a PBR material is held by a single texture called diffuse map. 
Diffuse maps are usually hand-made by artists or ``baked'' within external software. As a matter of fact, generic renderers compute function $r$ of Eq.~(\ref{eq:rendering}) by means of non-differentiable operations.


In the last years, the scientific community focused on alternative tools to implement a rendering function. In particular, researchers studied neural models to learn Eq.~(\ref{eq:rendering}) from data \cite{Kato_Ushiku_Harada_2017,mildenhall2020nerf,rematas2020neural}, or, more specifically, they promoted new rendering software that allows the user to compute gradients with respect to several parameters involved in the rendering process \cite{Liu_2019_ICCV,nimier2019mitsuba,ravi2020pytorch3d}. Among the latter category, we mention PyTorch3D \cite{ravi2020pytorch3d}, that implement a differentiable rendering API based on the widely diffused machine learning framework PyTorch.\footnote{See \url{https://pytorch.org/} and \url{https://pytorch3d.org/}} Despite being extremely versatile, several  differentiable renderers do not support PBR \cite{Liu_2019_ICCV,ravi2020pytorch3d} or other advanced rendering facilities, thus not reaching the level of photorealism that is typical of high-end non-differentiable renderers.

\subsection{Adversarial Objects}
\label{sec:adv}
The growing diffusion of deep learning methods and applications in  real-life scenarios \cite{grigorescu2020survey} poses serious concerns on  their robustness. In particular,  the vulnerability of their prediction performances to intentionally designed alterations of input data, i.e.,  adversarial examples \cite{biggio2013evasion,biggio2018wild,szegedy2013intriguing}, has been proven using several methods, such as Fast Gradient Sign Method (FGSM)~\cite{goodfellow2014explaining}, Projected Gradient Descent (PGD)~\cite{madry2017towards} and many others \cite{akhtar2018threat}.


Let us consider a classification task and a generic annotated pair $(x,y)$, where $x \in \mathbb{R}^d$ denotes an input pattern  and $y$ is the associated supervision. We also consider a neural network classifier $\mathcal{C(\cdot|\theta)}$ with parameters $\theta \in \mathbb{R}^p$. Let us indicate with $\bar{y}$ the output yielded by the classifier when processing $x$, $\bar{y} = \mathcal{C}(x|\theta)$, and the loss function $L(\bar{y}, y, x)$ that measures the mismatch between the prediction and the ground truth.
Neural classifiers have been proved to be vulnerable to the injection of adversarial perturbations in the input space, resulting in the misclassification of the pattern at-hand. 
In particular, an adversarial input $x + \delta$ causes $C(\cdot|\theta)$ to make wrong predictions, i.e.,  $\hat{y} = \mathcal{C}(x + \delta|\theta)$ with  $\hat{y} \neq y$.
In order to inject a perturbation $\delta$ that can be considered imperceptible to humans, a set of admissible perturbations $\mathcal{P}$ is defined. 
A common choice, that we will consider henceforth, limits the perturbation to fall upon the $\ell_{\infty}$-ball.
In the most simple case, the goal of the attacker is to find $\delta$ as solution of the following optimization problem,
\begin{equation}
   \max_{\delta \in \mathcal{P}  }    L(\bar{y}, y,x + \delta).
   \label{eq:opt_max}
\end{equation}
In the specific case of PGD, problem (\ref{eq:opt_max}) is solved by an iterative scheme, eventually including random restarts,
\begin{equation}
   {x}^{t+1} = \Pi_{\mathcal{P}} \big( {x}^{t}    + \alpha \cdot \texttt{sign}\big( (\nabla_{x} L)(\bar{y}, y,{x}^{t}\big)  \big)
\end{equation}
being ${x}^0 = x$,\footnote{Or ${x}^0 = x + \delta^0$, with  $\delta^0$ that is randomly generated.} $t$ the iteration index, $\alpha > 0$ the step length and $\Pi_{\mathcal{P}}$ projects its argument onto an $\ell_{\infty}$-ball with radius $\varepsilon$ centered on the original example.

In the case of 3D data, prior work~\cite{lu2017no,luo2015foveation} has shown that carefully-designed 2D adversarial examples fail to fool classifiers in the physical 3D world under several image transformations, such as changes in viewpoint, angle or other conditions (camera noise or light variation). 
In order to generalize attacks to such contexts, Athalye et al.~\cite{athalye2018synthesizing} proposed adversarial examples that are robust over a certain distribution of transformations. 
They introduced the so-called Expectation over Transformation (EOT), where Eq.~(\ref{eq:opt_max}) becomes
\begin{equation}
    \max_{\delta \in {\mathcal{P}_{\mathcal{E}'}}  }  \mathbb{E}_{t \sim \mathcal{T}} [L(\bar{y}, y,t(x + \delta))]
   \label{eq:opt_eot}
\end{equation}
being $\mathcal{T}$ a distribution of transformations and $t(\cdot)$ is a transformation sampled from $\mathcal{T}$, while $t(x + \delta)$ is the input of the classifier.
Moreover, ${\mathcal{P}_{\mathcal{E}'}}$ is the set of perturbations for which 
\begin{equation}
     \mathbb{E}_{t \sim \mathcal{T}}[d\big(t(x + \delta), t(x)\big)] < \varepsilon'
   \label{eq:delta_eot}
\end{equation}
where $d(\cdot,\cdot)$ is a distance function and $\varepsilon' > 0$.
This approach basically introduces expectations both in the objective function and in the perturbation-related constraint.
What is important to consider is that $\mathcal{T}$ is about a wide variety of transformations, including special operations that consist in using $x + \delta$ as a texture of a 3D object and rendering it to a 2D image.
The authors of~\cite{athalye2018synthesizing} use this intuition to physically create real-world 3D objects that are adversarial over different visual poses. Of course, this requires a renderer (Section~\ref{sec:rendering}) that is differentiable, and \cite{athalye2018synthesizing} is based on specific ad-hoc operations that cannot be easily implemented in a general setting.

Other very recent works focus on different aspects of the rendering process. In \cite{zeng2019adversarial}, authors consider the tasks of Visual Question Answering (VQA) and 3D shape classification, and they perturb multiple physical parameters such as the material, the illumination or the normal map. However, they keep a fixed view of the rendered object, that might create artifacts when the object is observed from different locations.
Liu et al.~\cite{liu2018beyond} proposed perturbations that are focussed on lighting. Their work is based on a ad-hoc created physically-based differentiable renderer capable to backpropagate gradients to the parametric space. 
MeshAdv~\cite{xiao2019meshadv} alters the object meshes, using a neural renderer applied on models  with constant reflectance -- very simple textures. The authors investigate the robustness under various viewpoints and the transferability to black-box renderers under controlled rendering parameters.
A recent work by Yao et al.~\cite{yao2020multiview} leverages 
multi-view attacks inspired by EOT, in order to devise 3D adversarial objects  perturbing the texture space, investigating the attack quality using multiple classifiers.
Finally, Liu et al.~\cite{liu2020spatiotemporal} considers the case of embodied agents performing navigation and question answering.
To better attack the task at hand, the perturbations are focused on the salient stimuli characterizing the temporal trajectory followed by the embodied agent to complete its task.


%

\def\salth{\tau_S}

\section{Adversarial Attacks to 3D Virtual Environments}
\label{sec:transf_3d_obj}

We consider the problem of generating adversarial 3D objects in the context of the 3D Virtual Environments of Section~\ref{sec:3dv}. The existing experiences in crafting 3D adversarial objects (Section~\ref{sec:adv}) have shown that it is indeed possible to create attacks that fool the classifier of a rendered scene.
However, existing works are strongly based on ad-hoc solutions, sometimes using specifically created renderers, limiting the attacks to a single view or considering very simple textures. 
They usually assume that the attacker has access to low-level properties of the renderer, such as the mapping of the view-space coordinates to the texture-space coordinates, or that renderers can be modified to expose additional information~\cite{athalye2018synthesizing}. Unfortunately, all these assumptions does not make their findings easily adaptable to more general cases. Another remarkable limit is that rendering engines (Section~\ref{sec:rendering}) are pieces of software that requires advanced skills not only in programming, but also in computer graphics, in order to be modified to accommodate attack procedures, or they might not be open source.

\begin{figure*}
        \centering
        \includegraphics[width=0.75\textwidth]{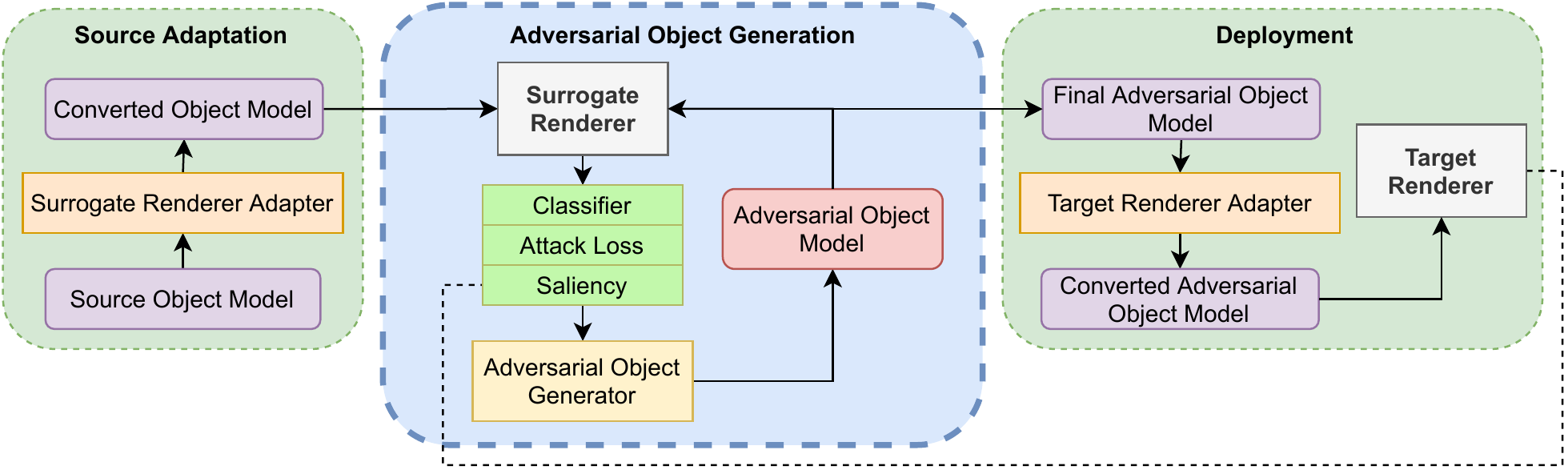}
        \caption{Structure of our adversarial object generation procedure. Saliency is computed exploiting the target renderer, as highlighted by the dotted line. }
        \label{fig:toolchain}
    \end{figure*}

We focus on a more generic perspective, that is based on a realistic setting in which the attacker has the goal of creating adversarial objects for a certain \textit{target renderer} on which he has limited control. We assume  attacker to have some skills in Adversarial Machine Learning but not necessarily an advanced knowledge of computer graphics. We explore the idea of synthesizing 3D adversarial objects using off-the-shelf popular software packages, well assembled into a specifically designed tool chain, with the goal of being able to craft malicious examples that can then be transferred to the target renderer of the considered 3D Virtual Environment. We report the structure of the proposed tool chain in Fig.~\ref{fig:toolchain}.

Our computational pipeline takes into account two different renderers (Fig.~\ref{fig:toolchain}, white boxes). One of them is the already introduced target renderer, while the other one is what we refer to as \textit{surrogate renderer}. The latter is a differentiable renderer on which the attacker has complete control, a reasonable assumption considering that many open-source differentiable renderers have been recently made available to the research community. As discussed in Section~\ref{sec:rendering}, it is likely that differentiable renderers will not perfectly match the quality of the target renderer, so that we focus on the specific case in which there is an evident difference between the outcome of the two renderers.
Of course, we assume that the 3D Virtual Environment allows users to  introduce and render custom objects.
However, care must be taken in adapting the object data format between the two renderers, since there might be a misalignment between the type of models expected by the 3D Virtual Environment and by the surrogate engine, requiring specific adaptations (Fig.~\ref{fig:toolchain}, leftmost and rightmost blocks).

Let us consider a certain object $o$ of class $y$, a scene $s$ and a camera $c$. 
The notation $o$ indicates the object and all its properties (mesh, textures, etc.), and, for the sake of simplicity, we indicate with $o + \delta$ an alteration of the object obtained by perturbing its properties by an offset $\delta$.
We consider the image (view) $I_{s,c,o}$ that we get when plugging $o$ into scene $s$, and rendering the whole 3D data when observed from camera $c$. We overload the notation of $r$ in Eq.~(\ref{eq:rendering}) to introduce the dependence on $o$,
\begin{equation}
    I_{s,c,o} = r(s,c,o).
\end{equation}
A neural network classifier $\mathcal{C}$ (Fig.~\ref{fig:toolchain}, top green box) processes $I_{s,c,o}$. 
The classifier prediction $\overline{y} = \mathcal{C}({I}_{s,c,o}|\theta)$ is evaluated into a loss function (Fig.~\ref{fig:toolchain}, mid green box) that drives the generation of the adversarial object, inspired by the EOT of Eq.~(\ref{eq:opt_eot}), even if fully focused on transformations in the 3D world. In particular,
\begin{equation}
    \max_{\delta \in \mathcal{P}}  \mathbb{E}_{(s,c) \sim \mathcal{S}} [L(\bar{y}, y, {I}_{s,c,o + \delta})]
   \label{eq:opt_our}
\end{equation}
where $\mathcal{S}$ includes different camera positions and orientations, different lighting conditions and, in the most generic cases, different backgrounds. Eq.~(\ref{eq:opt_our}) is paired with a norm-based constraint that ensures $\| I_{s,c,o} - {I}_{s,c,o+\delta}\|_{\infty} < \varepsilon$, for all $s,c$. This view-based constraint acts as an indirect measure to ensure that ${o}$ is not changing in a too evident way. 

    
In order to solve Eq.~(\ref{eq:opt_our}) we need to exploit a differentiable renderer that allows us to compute the gradient with respect to the properties of object $o$. 
In general, we do not know in advance how many of such properties will be concretely altered by $\delta$, since it depends on what transformations are considered and on the details of the experimental setup. Of course, perturbing a small subset of such properties could be a desirable feature to ensure that the object is not altered in a too evident way. 
Inspired by this consideration, we propose an approach that is agnostic to the type of adversarial example generation algorithm. In particular, we exploit the fact that while the target renderer cannot be used to compute gradients with respect to $o$, we can indeed use it to render an object view, that we indicate with $I^{T}_{s,c,o}$ for a certain pair $(s,c)$. Then, we can straightforwardly compute the gradients of $L(\bar{y},y, I^T_{s,c,o})$ with respect to the pixels of the image. 
The pixels with the largest (absolute) gradients are the ones to which the classifier is more sensitive. Once we rescale such gradients in $[0,1]$, we can select a custom threshold $\salth \in (0,1)$ to build a binary \textit{saliency map} (Fig.~\ref{fig:toolchain}, bottom green box) that tells what are the pixels to which the \textit{target} renderer is known to be significantly sensitive. Moving to the \textit{surrogate} renderer, such map can be used to avoid gradients to be back-propagated through not salient pixels, that, in turn, is expected to reduce the number of properties of $o$ that will be altered.



\subsection{Case Study}
\label{sec:case_study}

\begin{figure}
        \centering
         \includegraphics[width=0.9\linewidth]{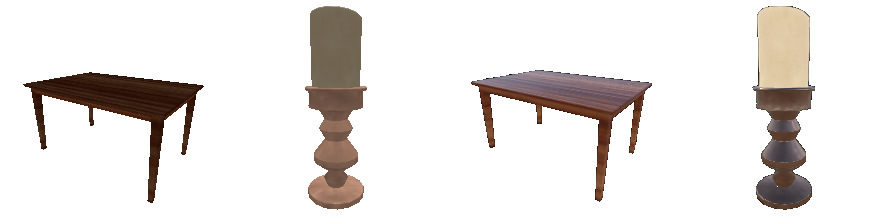}\\
                  \hspace{0.8cm} (a) PyTorch3D \hspace{1.3cm} (b) SAILenv (Unity3D)\\
        \caption{Rendering capabilities of the surrogate (a) and target (b) renderers.}
        \label{fig:rendering_comparison}
    \end{figure}

We instantiated the strategy of Fig.~\ref{fig:toolchain} into a specific case study, that will also drive our experiments.
Our choices are completely driven by simplicity, selecting tools that are recent, freely available, and that do not require advanced skills in computer graphics.
In particular, we considered a recent 3D Virtual Environment named SAILenv~\cite{Meloni2020} (Section~\ref{sec:3dv}), that is open-source and claimed to be simple to be interfaced with Machine Learning models.
SAILenv exploits Unity3D, that is our target renderer, which satisfies the assumption of allowing the attacker to render custom objects while having limited low-level control to the renderer, since it is based on proprietary code and cannot be ``easily'' modified. As  surrogate renderer we focused on the recent PyTorch3D~\cite{ravi2020pytorch3d} (Section~\ref{sec:rendering}), that is completely based on Autograd and thus trivial to integrate with a PyTorch-based classifier for gradient computation.
The two renderers
have some remarkable differences. SAILenv uses PBR while PyTorch3D is based on diffuse-based rendering, which takes into account only the surface color and a much simplified light model. Both are discussed in Section~\ref{sec:rendering}.
We approximate the diffuse maps rendered by Pytorch3D with the Albedo Texture Maps used within Unity3D. This approximation holds the best for neutral illumination settings and for low reflective materials.
See Fig.~\ref{fig:rendering_comparison} for a comparison of the rendering capabilities of the two renderers. 
PyTorch3D allows gradients estimation of several parameters of the object and of the scene (surface color, object geometry, lightning, etc.). For the scope of this paper, we will focus only on the surface color texture. 
This is a very challenging setting due to the aforementioned limited rendering facilities of PyTorch3D, making this case study a very good representative of the previously described attack scenario.

\begin{figure}
    \centering
    \includegraphics[width=0.15\textwidth]{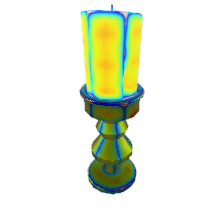}
    \includegraphics[width=0.15\textwidth]{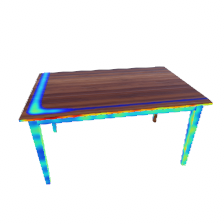}
    \includegraphics[width=0.15\textwidth]{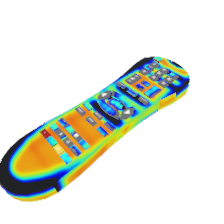}
    \caption{Multi-view saliency maps (target renderer), projected into the surface of the object (projection computed by the surrogate engine) -- red=high; blue=low.}
    \label{fig:saliency_maps}
\end{figure}

We qualitatively show in Fig.~\ref{fig:saliency_maps} how the saliency maps, computed using Unity3D over multiple views, are projected back onto the texture space, accumulating their contributes on the texels, then rendering the 3D objects. While computing this projection in the target renderer is not straightforward, this can be easily done following the texturing routine of PyTorch3D, and that is how we created the figure. We can appreciate how the larger saliency areas only cover a subportion of the texels.
Notice that the data adapters or Fig.~\ref{fig:toolchain} (i.e.\ Surrogate Renderer Adapter and Target Renderer Adapter) play a crucial role in our case study, since Unity3D stores objects in a different format (FBX) than the one used by PyTorch3D (OBJ). 
We implemented a source object converter by means of a  Blender\footnote{\url{https://www.blender.org/}}-based script, created from scratch. The final adversarial object is then converted back to the Unity3D format through a plugin that is internal to Unity3D, and finally rendered in SAILenv.

\begin{figure*}
        \centering
            \rotatebox{90}{\footnotesize{PyTorch3D}}
             \includegraphics[width=0.95\linewidth]{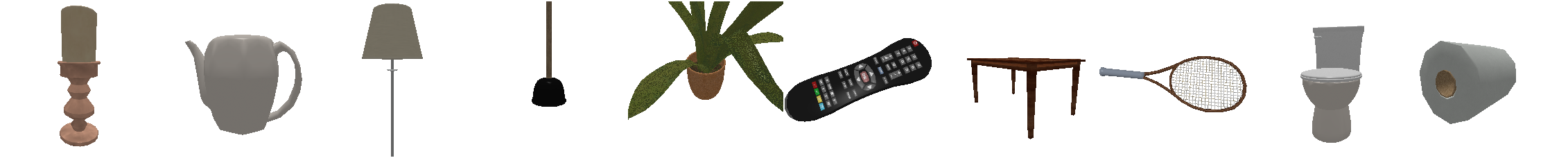}
             
            \rotatebox{90}{\hskip 3mm \footnotesize{SAILenv}}
             \includegraphics[width=0.95\linewidth]{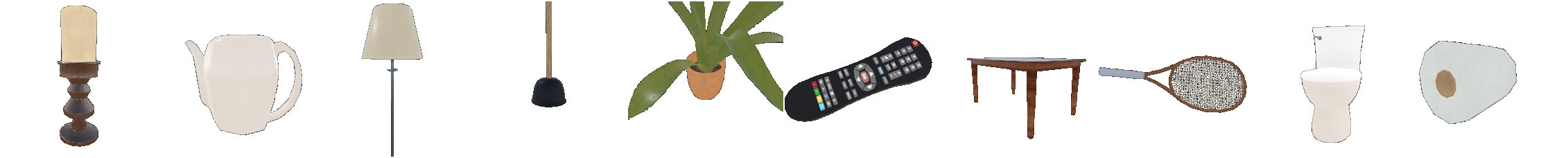}
        \caption{Objects considered in our case study, rendered using  PyTorch3D (top) and in SAILenv/Unity3D (bottom).}
        \label{fig:objects_grid}
    \end{figure*}

In order to evaluate the impact of our strategies in different networks, we selected two popular and powerful deep neural image classifiers trained on ImageNet, that are \textsc{InceptionV3} and \textsc{MobileNetV2}.\footnote{\url{https://pytorch.org/hub/pytorch_vision_mobilenet_v2/}\\ \url{https://pytorch.org/hub/pytorch_vision_inception_v3/}} The former is a state-of-the art image classifier, the latter is a smaller model, still very accurate. 
We considered 10 different objects from the SAILenv library, associated to classes that are supported by the classifiers. 
The objects are shown in Fig.~\ref{fig:objects_grid}, comparing their appearance in the surrogate and target renderers.
As adversarial object generation method, we implemented the PGD attack described in Section~\ref{sec:adv}, using the cross entropy loss.


\section{Experimental Results}
\def\npixels{N_\%}
\def\accdrop{A_{drop}}
\def\accbef{A_{before}}
\def\accaft{A_{after}}
\def\incept{\textsc{InceptionV3}} 
\def\mobnet{\textsc{MobileNetV2}}
We performed several experiments to evaluate the proposed attack strategy in the case study of Section~\ref{sec:case_study}.\footnote{Our implementation of what we propose and study in this paper can be found at \url{https://github.com/sailab-code/SAIFooler}. The 3D models used in the experiments can be found at \url{http://sailab.diism.unisi.it/sailenv/} (download section).}
Each object is rendered from $60$ different views, keeping the camera at a fixed distance which was manually chosen to obtain an iconic image of the object, i.e., so that the object covers most of the picture. The camera turns around the object, from $0^{\circ}$ to $360^{\circ}$ and also changes its elevation. The range on which the elevation is changed is manually chosen for each object in order
to avoid unnatural viewing orientations that would lower the classification accuracy even without any attacks. We considered a directional light, similar to the way sunlight shines on objects, coming from the front and at an elevation of $75^{\circ}$. The background scene of each object is composed of a uniform color, that was evaluated as being white or black, selecting the one that maximized the recognition accuracy.

We explored attacks that progressively yield larger alterations in the original textures, considering $\varepsilon \in \{0.05, 0.1, 0.5\}$, comparing cases in which we do not use saliency maps or when the maps are binarized with different thresholds of tolerance, i.e., $\tau_S \in \{0.05, 0.2\}$, and we set $\alpha$ to $0.01$.        
It is important to remark that we are considering the $\ell_{\infty}$ norm to bound the perturbations, so that, given the same $\varepsilon$, we can have very different number of altered texels. Altering less texels is expected to reduce the probability of letting humans recognize the adversarial object, and that is the goal of the proposed saliency-map-based procedure.
We used two metrics to evaluate the quality of the adversarial objects. The first one is the \textit{accuracy drop} $\accdrop$, that is the ratio of the variation of average accuracy (before and after the attack, referred to as  $\accbef$ and $\accaft$, respectively) to the initial accuracy, while the second one is the percentage of texels $\npixels$ that are altered by the attack procedure. Formally,
\begin{eqnarray*}
\accdrop &=& \frac{\accbef{} - \accaft}{\accbef}\\
\npixels &=& \frac{\|\texttt{tex}_{before}-\texttt{tex}_{after}\|_1}{ |\texttt{tex}_{before}|},
\end{eqnarray*}
being $\texttt{tex}_{\cdot}$ the texture tensor composed of $|\texttt{tex}_{before}|$ elements and $\|\cdot\|_1$ the $\ell_1$ norm.       
We computed both the metrics within the PyTorch3D renderer and the SAILenv (Unity3D) renderer. In the former case, we are basically exploring a white-box scenario, where the system we attack is the one on which we evaluate the result. In the latter case, we consider the impact of the adversarial object once it is transferred to a target environment, in a very challenging black-box setting, due to the previously described differences between the two renderers.


        
        


In Table \ref{tab:acc_drop} we report the main results of our experiments, showing $\accdrop$ for the considered objects and the average result (last column; we indicate with \textit{n.a.} those objects that were not correctly recognized by \mobnet\ in their original state). 
In the case of the surrogate renderer, it is evident that even lower values of $\varepsilon$ are enough to usually achieve near $100\%$ drop of accuracy, with the exception of the Tennis Racket, for which a higher $\epsilon$ is needed. 
\begin{table*}
    \def\figsize{0.07\textwidth}
    \centering
        \caption{Accuracy drop in each considered object and average result.}
    \hskip 1mm
     \resizebox{1.0\textwidth}{!}{\begin{tabular}{cl|c|cccccccccc|c}
        \toprule
               &   & $\varepsilon$ &  \includegraphics[width=\figsize]{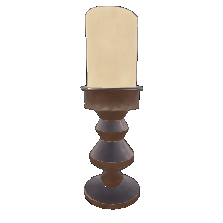} &  \includegraphics[width=\figsize]{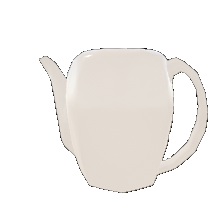} &  \includegraphics[width=\figsize]{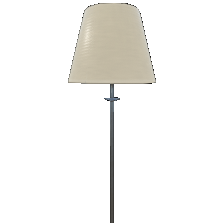} &  \includegraphics[width=\figsize]{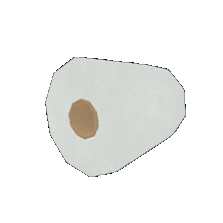} &  \includegraphics[width=\figsize]{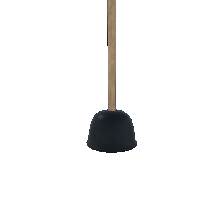} &   \includegraphics[width=\figsize]{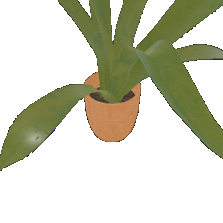} &  \includegraphics[width=\figsize]{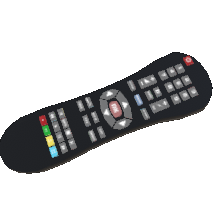} &  \includegraphics[width=\figsize]{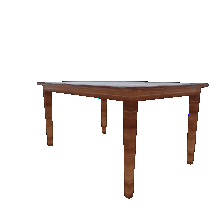} &  \includegraphics[width=\figsize]{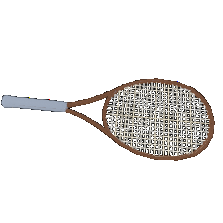} &  \includegraphics[width=\figsize]{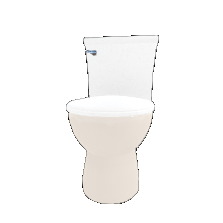} &   \textbf{Avg} \\

        \midrule
        \multirow{6}{*}{\rotatebox{90}{PyTorch3D}} 
        &    \multirow{3}{*}{\scriptsize \mobnet} & 0.05 &     1.00 &   n.a. &        0.86 &           1.00 &     1.00 &  1.00 &               0.98 &               0.95 &           0.92 &     1.00 &  \textbf{0.97} \\
     &             & 0.10 &     1.00 &   n.a. &        0.86 &           1.00 &     1.00 &  1.00 &               0.98 &               1.00 &           1.00 &     1.00 &  \textbf{0.98} \\
      &            & 0.50 &     1.00 &   n.a. &        0.86 &           1.00 &     1.00 &  1.00 &               1.00 &               0.95 &           1.00 &     1.00 &  \textbf{0.98} \\
      
        \cmidrule{2-14}
        & \multirow{3}{*}{\scriptsize \incept} & 0.05 &     1.00 &   1.00 &        0.98 &           1.00 &     1.00 &  0.97 &               0.95 &               1.00 &           0.05 &     1.00 &  \textbf{0.89} \\
        &  & 0.10 &     1.00 &   1.00 &        0.95 &           1.00 &     0.98 &  0.97 &               1.00 &               0.97 &           0.45 &     1.00 &  \textbf{0.93} \\
        &  & 0.50 &     1.00 &   1.00 &        1.00 &           1.00 &     1.00 &  0.94 &               0.97 &               1.00 &           0.97 &     1.00 &  \textbf{0.99} \\
    
        \toprule




    \multirow{6}{*}{\rotatebox{90}{SAILenv}}  &  \multirow{3}{*}{\scriptsize \mobnet} & 0.05 &    0.76 &   n.a. &        0.62 &          1.00 &     0.72 &  0.60 &               0.21 &               0.00 &           0.81 &     n.a. &  \textbf{0.59} \\
       &           & 0.10 &    0.72 &   n.a. &        0.62 &          1.00 &     0.74 &  0.65 &               0.43 &               0.00 &           0.90 &     n.a. &  \textbf{0.63} \\
       &           & 0.50 &    0.76 &   n.a. &        0.62 &          1.00 &     0.74 &  0.68 &               0.41 &               0.00 &           0.94 &     n.a. &  \textbf{0.64} \\
            \cmidrule{2-14}
      
       &  \multirow{3}{*}{\scriptsize \incept} & 0.05 &    0.37 &  0.87 &       -0.12 &          0.65 &     0.07 &  0.00 &               0.10 &               0.16 &           0.00 &     1.00 &  \textbf{0.31} \\
         &     & 0.10 &    0.41 &  0.77 &       -0.15 &          0.65 &     0.08 &  0.21 &               0.34 &               0.47 &           0.00 &     1.00 &  \textbf{0.38} \\
       &       & 0.50 &    0.39 &  0.73 &       -0.08 &          0.65 &     0.12 &  0.09 &               0.41 &               0.42 &           0.07 &     1.00 &  \textbf{0.38} \\
        \bottomrule
        \end{tabular}}
        \label{tab:acc_drop}
\end{table*}            
When the attack is transferred to SAILenv, we can observe that the classifiers are still fooled in a non-negligible manner. Of course, the extent to which the attack has effect is reduced, as expected, but it is surprisingly to see that even if the difference between the two renderers in our case study is significant, the attack can impact the outcome of the classification in the target 3D Virtual Environment. 
With the exception of Lamp Floor, Pot, Tennis Racket in the case of \incept, and Teapot, Living Room Table for \mobnet, where the attack yields no evident accuracy drops (in one case also a negative drop, meaning that it is slightly improving the classification), the other adversarial objects reduce the accuracy of the classifiers, with a pretty strong effect in the case of \mobnet.



In Fig.~\ref{fig:scatterplots}, we report the 2D plot of $\accdrop$ against $\npixels$ (all objects), taking into account different values of $\varepsilon$ and saliency thresholds $\salth$.
When using PyTorch3D, several points are clustered on the right side of the plot, associated to a large $\accdrop$. In the case of SAILenv and \incept\ as a classifier, the majority of points are  between $0.1$ and $0.5$, with some attacks reaching very large drops.
In the case of SAILenv and \mobnet, more attacks have $\accdrop$ approaching $1.0$. 
As already discussed, even small $\varepsilon$ might end up in altering a significant amount of texels. However, the plots show that using saliency is a good solution to identify a trade off between $\accdrop$ and $\npixels$. In particular, the attacks in which no saliency information is used are usually located in the upper-right quadrant of the plot -- high impact but they heavily alter the textures. Attacks with the largest saliency threshold $\salth$ are instead usually located in the lower-left quadrant -- low impact but they are also more hardly noticeable by humans, altering less pixels. When using a lower $\salth$ we get results distributed in the central part of the plot -- good impact on the classifier, altering a relatively small number of texels.
                \def\scattersize{0.23\textwidth}
                \begin{figure}
                    \centering
                    
                         \scriptsize{\hspace{0.5cm}PyTorch3D \hspace{2.6cm} SAILenv}
            
                        \scriptsize{\rotatebox{90}{\hspace{1.4cm}MobileNet}}
                        {\centering
                        \includegraphics[width=\scattersize]{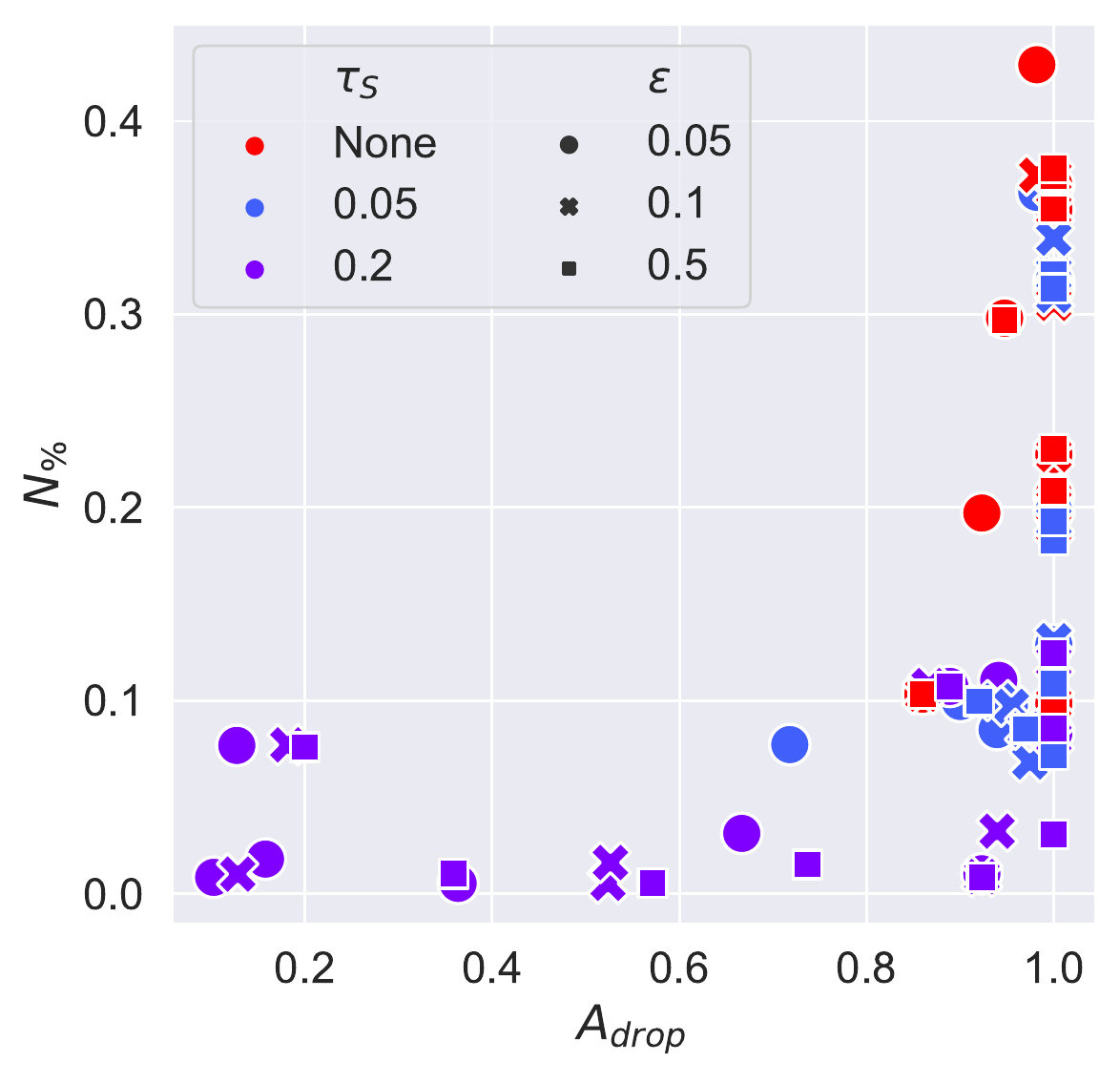}
                         \includegraphics[width=\scattersize]{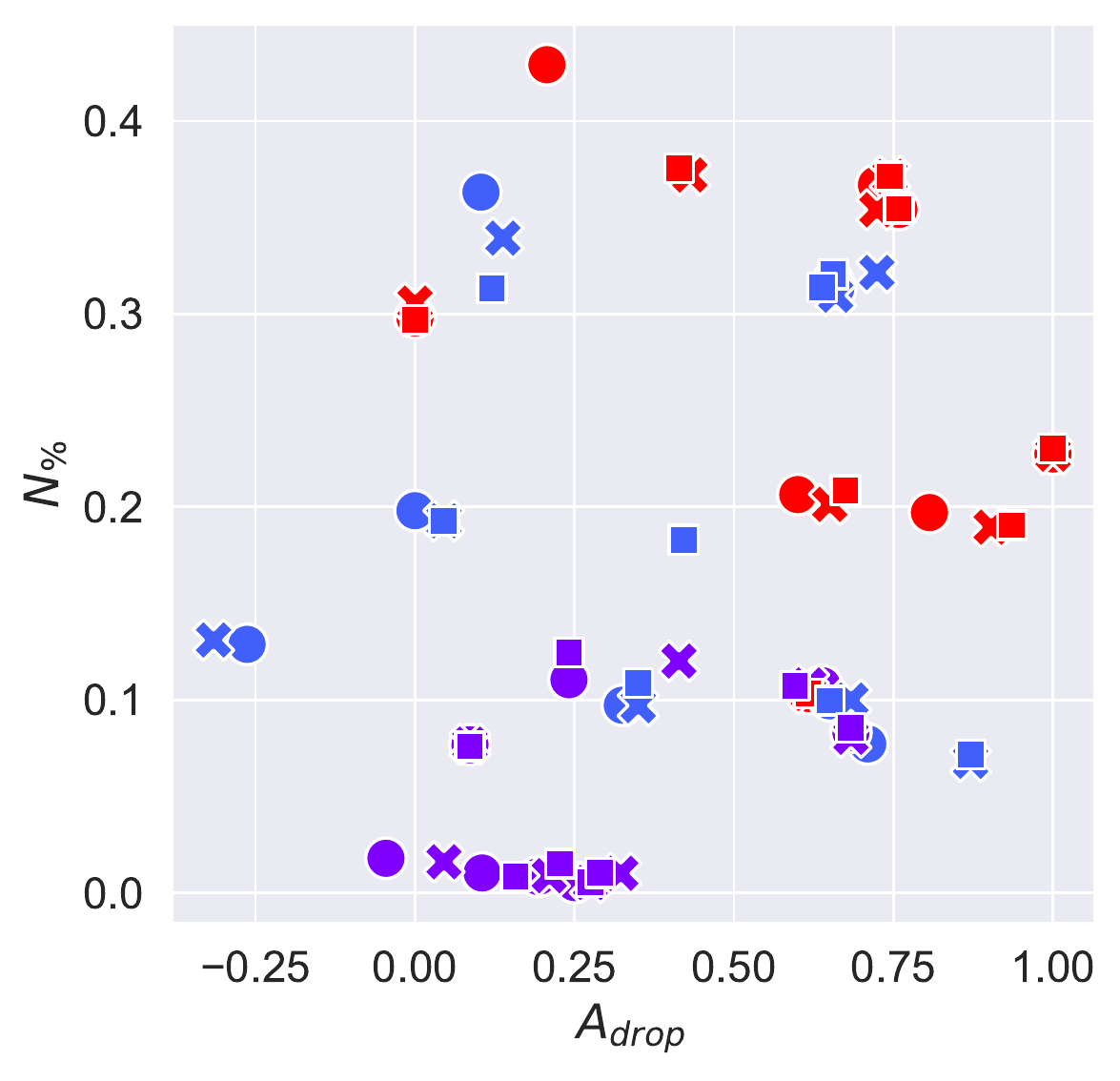}
                         }
                         
                         \scriptsize{\rotatebox{90}{\hspace{1.4cm}Inception}}
                         {\centering
                         \includegraphics[width=\scattersize]{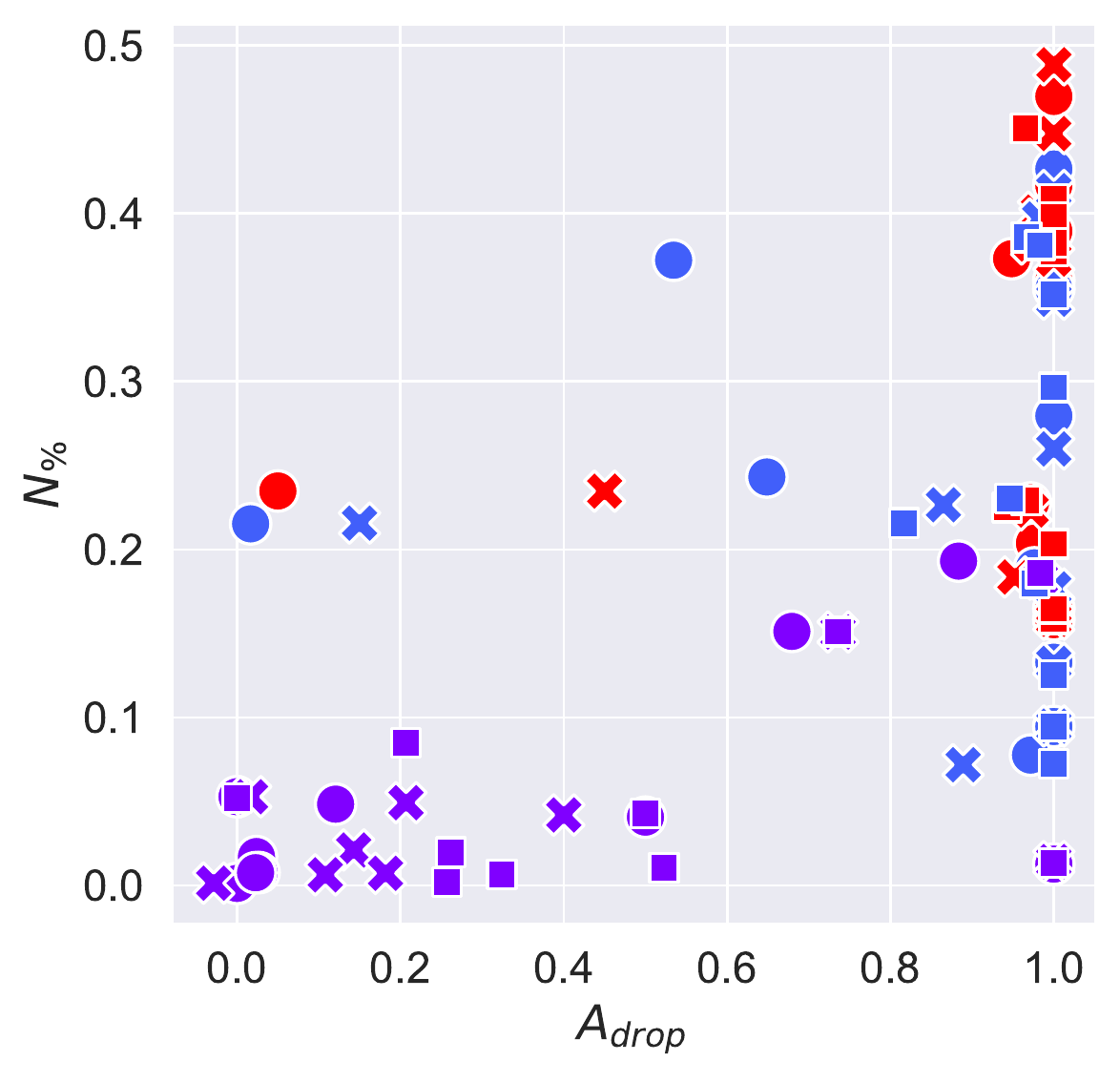}
                         \includegraphics[width=\scattersize]{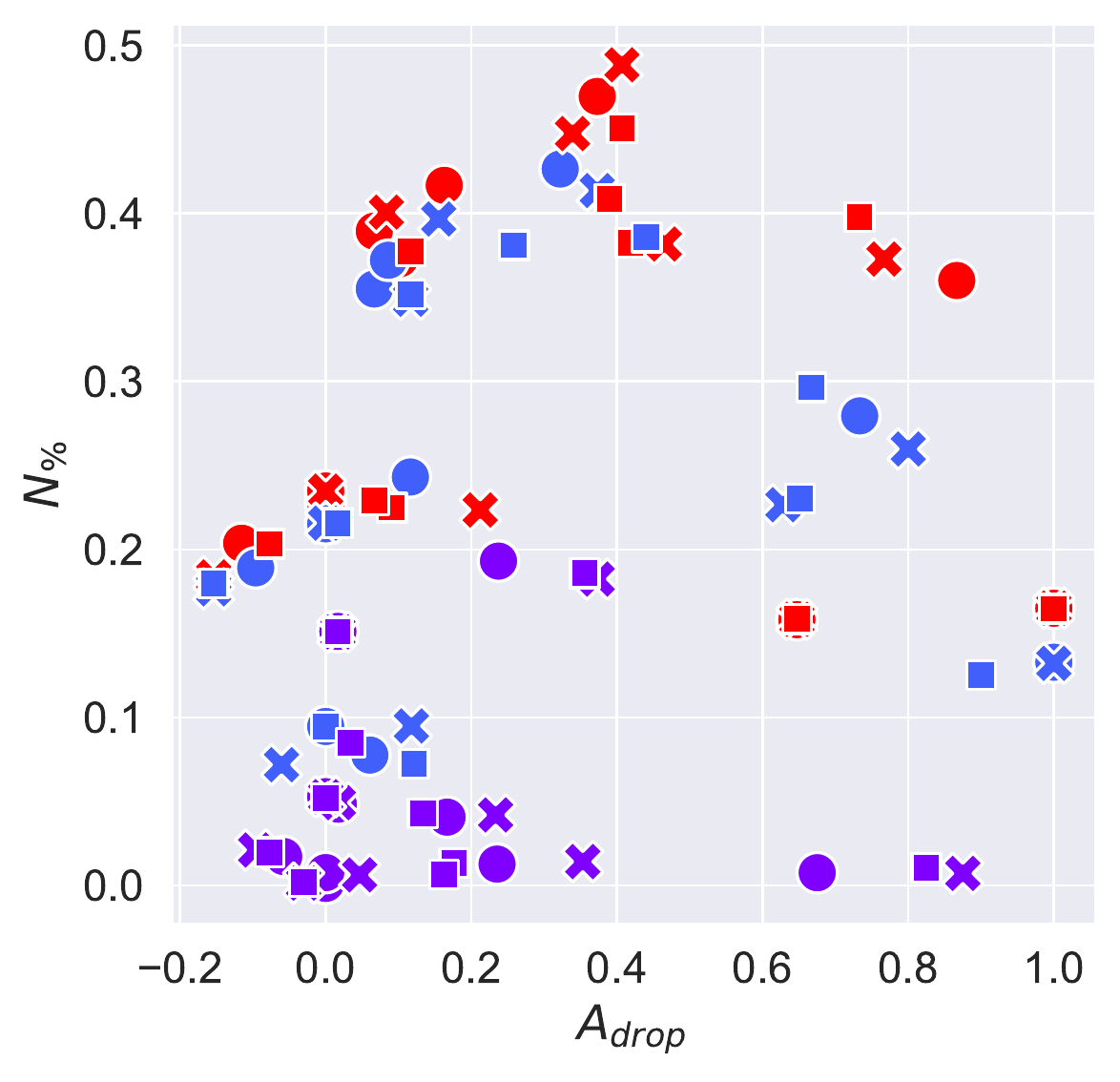}
                         }
                     
                     \caption{Accuracy drop versus percentage of altered texels. Points are about adversarial objects (colors indicate different $\salth$; markers are about different  $\varepsilon$).}
                    \label{fig:scatterplots}
                \end{figure}

\begin{figure}
                \centering
                 \includegraphics[width=0.23\textwidth]{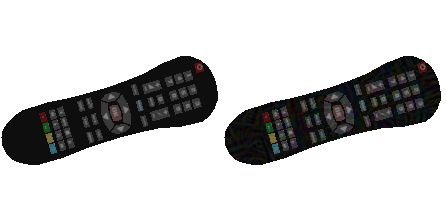}
                 \includegraphics[width=0.23\textwidth]{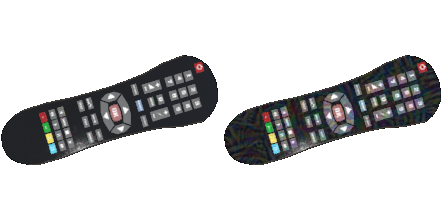}\\
                \includegraphics[width=0.23\textwidth]{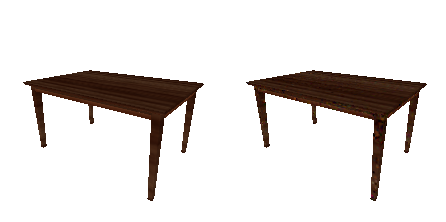}
                \includegraphics[width=0.23\textwidth]{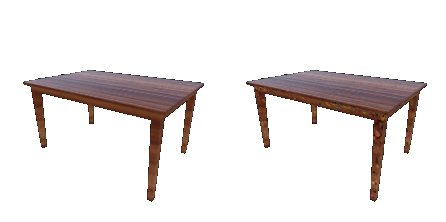} \\
                 \hspace{0.7cm} (a) PyTorch3D \hspace{1.3cm} (b) SAILenv (Unity3D)\\
             \caption{For the surrogate (a) and target (b) renderers, we report two objects (one per line), before (left) and after (right) the attack.}
            \label{fig:remote_predictions}
            \end{figure}

\begin{figure}[ht]
                \centering
                {PyTorch3D \hskip 2.0cm SAILenv}\\
                \rotatebox{90}{\hskip -2.2cm Remote Control}
                 \includegraphics[width=0.23\textwidth,trim=0 0 0 0,clip,valign=t]{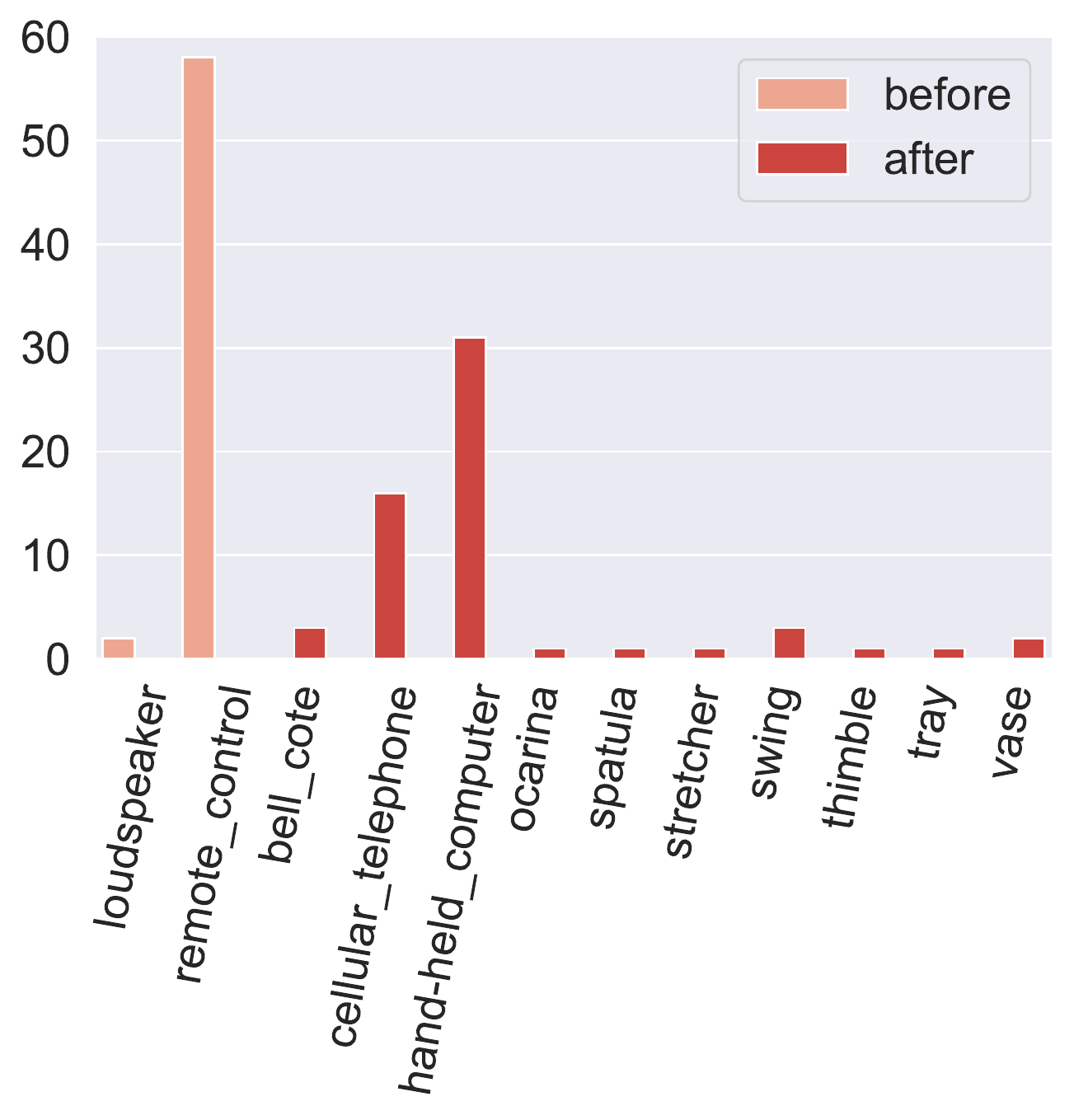}
                 \hskip -2mm
                 \includegraphics[width=0.23\textwidth,trim=0 0 0 0,clip,valign=t]{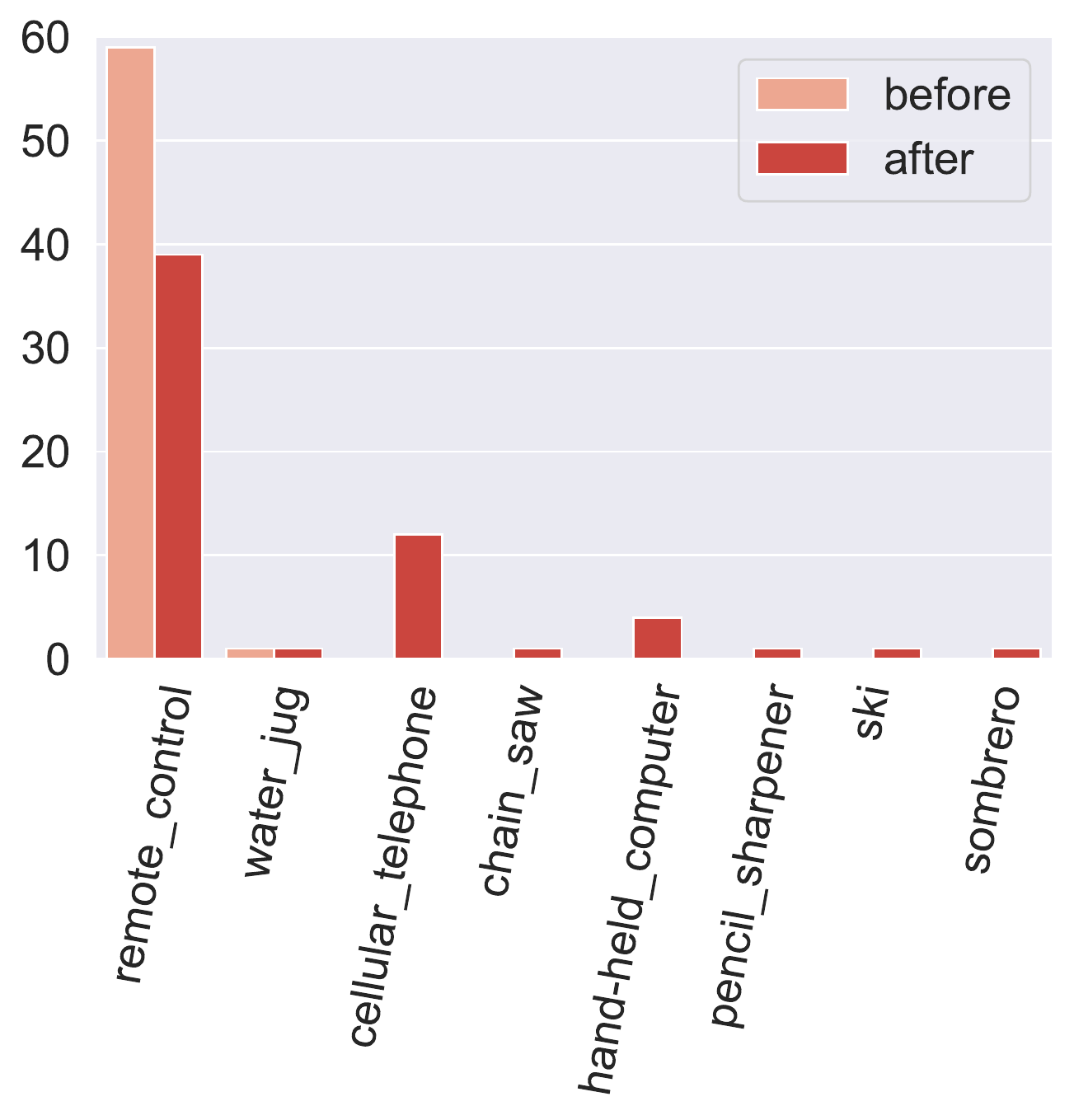}\\
                 {PyTorch3D \hskip 2.0cm SAILenv}\\
                 \rotatebox{90}{\hskip -2.0cm Dining Table}
\includegraphics[width=0.23\textwidth,trim=0 0 0 0,clip,valign=t]{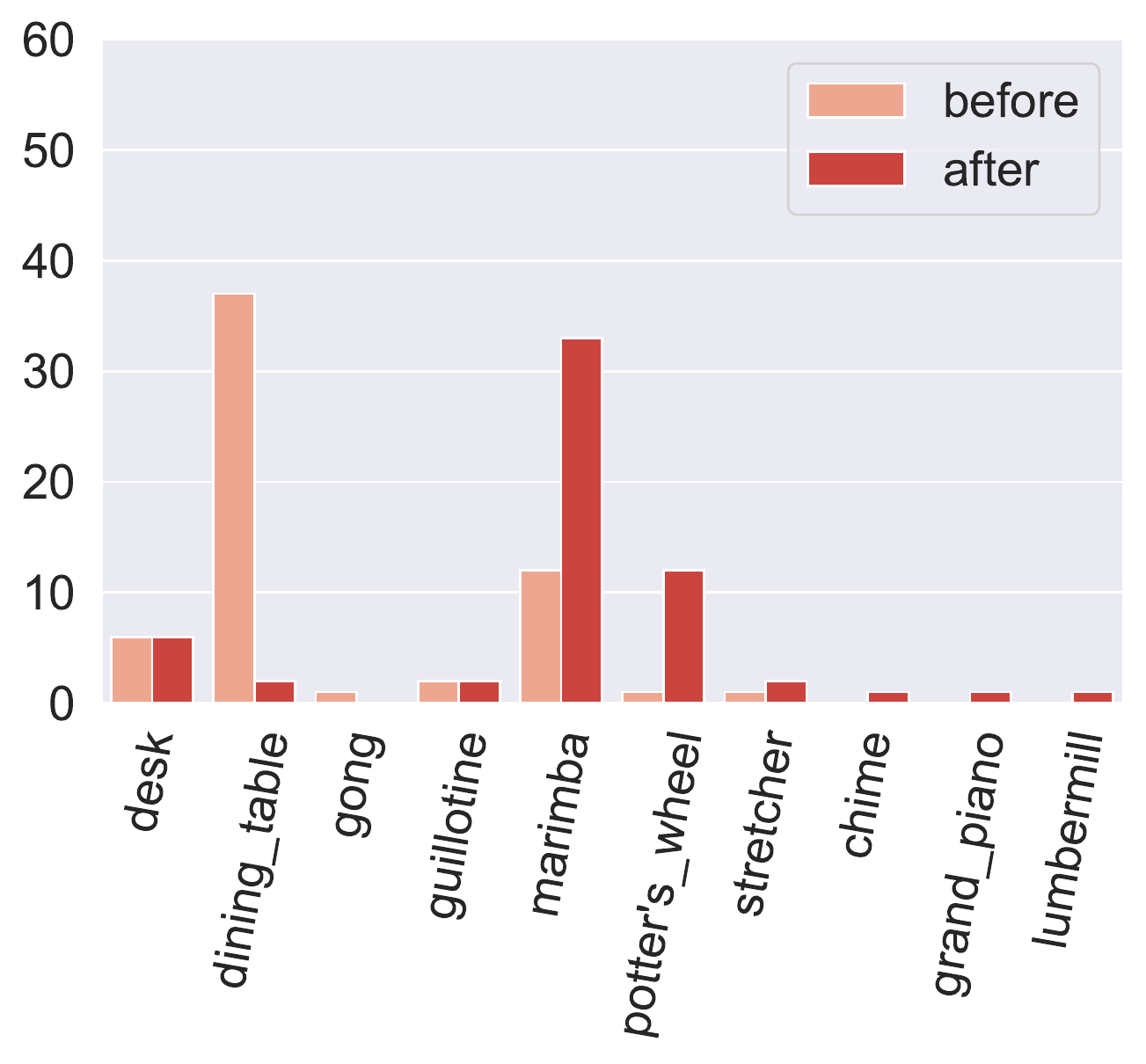}
        \hskip -2mm
         \includegraphics[width=0.23\textwidth,trim=0 0 0 0,clip,valign=t]{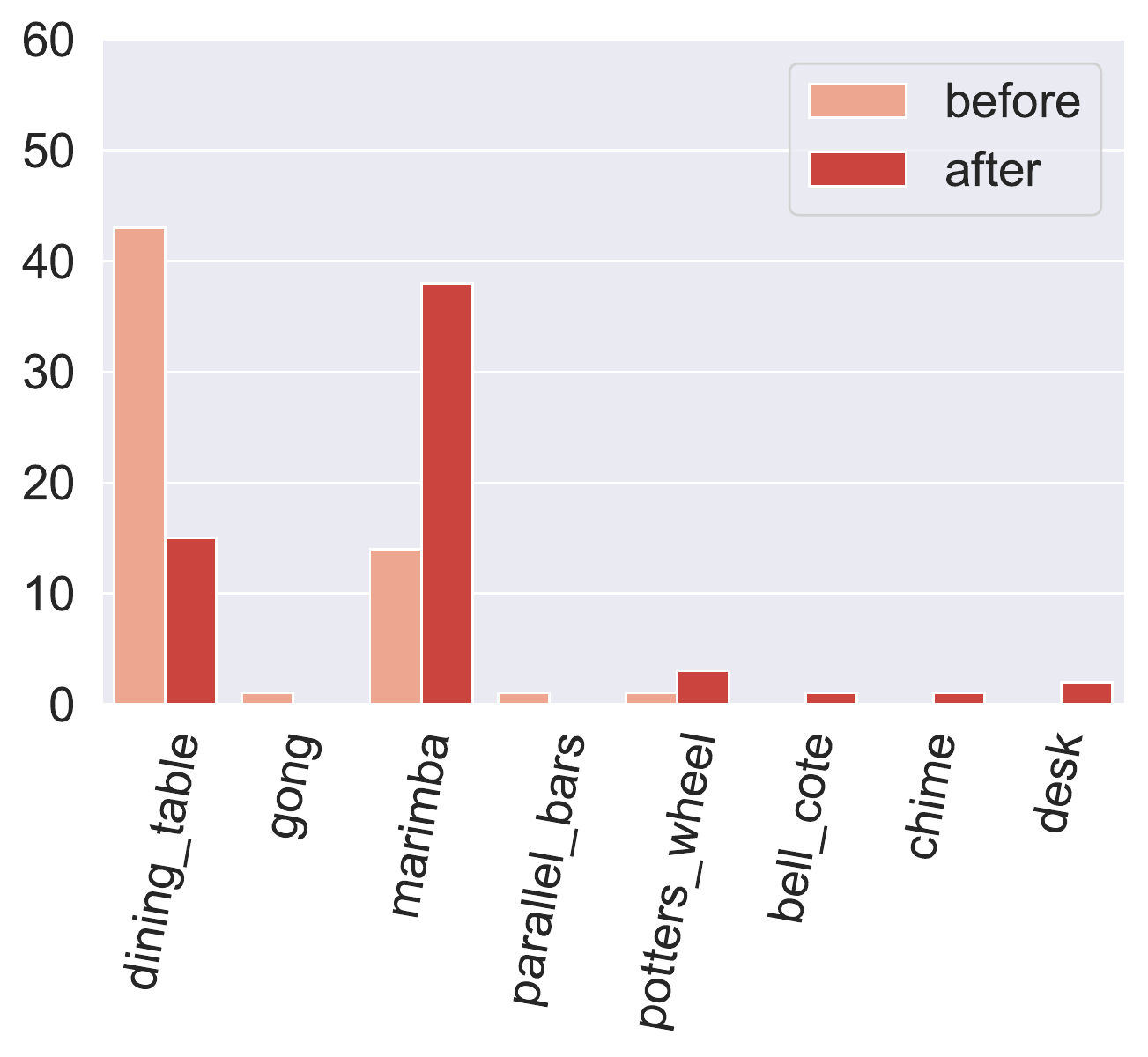}\\
             \caption{Number of correct predictions made by \incept\ out of 60 different views of the objects of Fig.~\ref{fig:remote_predictions}, before and after having attacked them.} 
            \label{fig:histo}
            \end{figure}

We qualitatively evaluated the renderings of the adversarial objects, reporting in Fig.~\ref{fig:remote_predictions} two examples that fool the \incept\ classifier in both the renderers. 
While we do not see any evident differences comparing the original and the altered objects, we still observe the huge gap between what the surrogate and the target renderer produce, remarking the importance of the results of this study.
In Fig.~\ref{fig:histo} we consider all the $60$ views of such objects, reporting how the predictions of \incept\ are distributed. It is evident that before the attack, most of the predictions are correctly distributed on the ground truth class, while after the attack they are spread over multiple incorrect classes. 
            

\section{Conclusions and Future Work}
\label{sec:conlcusions}
We presented a novel study on the transferability of adversarial 3D objects, created using an off-the-shelf differentiable renderer and then moved to a powerful 3D engine that is at the basis of several recent 3D Virtual Environments. Our analysis showed that it is indeed possible to setup a tool chain based on simple elements that do not require advanced skills in computer graphics, and use it to craft malicious 3D objects. Experiments on texture-oriented manipulations showed that attacks can be transferred to fool popular neural classifiers, also considering an estimated saliency of the texels. 
There is certainly room for future work in improving the effectiveness of the attacks (e.g., considering other parameters of the renderer -- mesh and others). However, our results are expected to point the attention of the scientific community towards this double-sided aspect: on one hand, it could be an issue for community-open 3D Virtual Environments, and, on the other hand, it is an opportunity to create even more powerful testing environments, purposely populated with adversarial examples.


%
%
%
%
\bibliographystyle{IEEEtran}
\bibliography{biblio}

\end{document}